# Online Bandit Learning against an Adaptive Adversary: from Regret to Policy Regret

**Raman Arora**     ARORA@TTIC.EDU
Toyota Technological Institute at Chicago, Chicago, IL 60637, USA

**Ofer Dekel**     OFERD@MICROSOFT.COM
Microsoft Research, 1 Microsoft Way, Redmond, WA 98052, USA

**Ambuj Tewari**     AMBUJ@CS.UTEXAS.EDU
Department of Computer Science, University of Texas, Austin, Texas 78712, USA

## Abstract

Online learning algorithms are designed to learn even when their input is generated by an adversary. The widely-accepted formal definition of an online algorithm's ability to learn is the game-theoretic notion of regret. We argue that the standard definition of regret becomes inadequate if the adversary is allowed to adapt to the online algorithm's actions. We define the alternative notion of *policy regret*, which attempts to provide a more meaningful way to measure an online algorithm's performance against adaptive adversaries. Focusing on the online bandit setting, we show that no bandit algorithm can guarantee a sublinear policy regret against an adaptive adversary with unbounded memory. On the other hand, if the adversary's memory is bounded, we present a general technique that converts any bandit algorithm with a sublinear regret bound into an algorithm with a sublinear policy regret bound. We extend this result to other variants of regret, such as switching regret, internal regret, and swap regret.

## 1. Introduction

Online learning with bandit feedback is commonly described as a repeated game between a player and an adversary. On each round of the game, the player chooses an action $X_t$ from an action set $\mathcal{X}$, the adversary chooses a loss function $f_t$, and the player suffers a loss of $f_t(X_t)$. We often assume that $f_t(X_t)$ is bounded in $[0,1]$. The player observes the loss value $f_t(X_t)$ and uses it to update its strategy for subsequent rounds. Unlike the full-information player, the bandit player does not observe the entire loss function $f_t$. The player's goal is to accumulate the smallest possible loss over $T$ rounds of play.

While this presentation is intuitively appealing, it hides the details on what information the adversary may use when choosing $f_t$. Since this aspect of the problem is the main focus of our paper, we opt for a less common, yet entirely equivalent, definition of the online bandit problem.

We think of online prediction with bandit feedback as an iterative process where only the player makes active choices on each round. The adversary, on the other hand, prepares his entire sequence of loss functions in advance. To ensure that this assumption does not weaken the adversary, we make the additional assumption that the loss function $f_t$ takes, as input, the player's entire sequence of past and current actions $(X_1, \ldots, X_t)$, which we abbreviate by $\mathbf{X}_{1,\ldots,t}$. More formally, for each $t$, $\mathcal{F}_t$ is a class of loss functions from $\mathcal{X}^t$ to the unit interval $[0,1]$, and the adversary chooses each $f_t$ from the respective class $\mathcal{F}_t$.

We model the player as a randomized algorithm that defines a distribution over $\mathcal{X}$ on each round and samples $X_t$ from this distribution. Therefore, even though $f_t$ is a deterministic function fixed in advance, the loss $f_t(\mathbf{X}_{1,\ldots,t})$ is a bounded random variable. The player observes the value of this random variable and nothing else, and uses this value to define a new distribution over the action space.

Interesting special cases of online learning with bandit feedback are the *k-armed bandit* (Robbins, 1952; Auer et al., 2002), *bandit convex optimization* (Kleinberg, 2004; Flaxman et al., 2005; Abernethy et al., 2008), and *bandit submodular minimization* (Hazan & Kale, 2009). In the $k$-armed bandit problem, the action space $\mathcal{X}$ is the discrete set $\{1, \ldots, k\}$ and each $\mathcal{F}_t$ contains all functions from $\mathcal{X}^t$ to $[0,1]$. In bandit convex optimization, $\mathcal{X}$ is a predefined convex set and each $\mathcal{F}_t$ is the set of functions that are convex in their last argument. A special case of bandit convex optimization is *bandit linear optimization* (Awerbuch & Kleinberg, 2004; Bartlett et al., 2008), where the functions in $\mathcal{F}_t$ are linear in their last argument. In bandit submodular minimization, $\mathcal{X}$ is the power-set of $\{1, \ldots, k\}$ and each $\mathcal{F}_t$ contains all of the functions that are submodular in their last argument.

Various different adversary types have been proposed in the literature (Borodin & El-Yaniv, 1998; Cesa-Bianchi & Lugosi, 2006). All adversary types are strategic and possibly malicious, have unlimited computational power, and are free to use random bits when choosing their loss functions. If the adversary is not restricted beyond the setting described above, he is called an *adaptive* adversary. Other adversary types are restricted in various ways. For example, an *oblivious* adversary is restricted to choose a sequence of loss functions such that each $f_t$ is oblivious to the first $t-1$ arguments in its input. In other words, $f_t$ can only be a function of the current action. Formally,

$$f_t(x_1, \ldots, x_t) = f_t(x'_1, \ldots, x'_{t-1}, x_t) ,$$

for all $x_1, \ldots, x_t$ and $x'_1, \ldots, x'_{t-1}$ in $\mathcal{X}$.

The *expected cumulative loss* suffered by the player after $T$ rounds (which we abbreviate simply as *loss*) is $\mathbb{E}\bigl[\sum_{t=1}^T f_t(\mathbf{X}_{1,\ldots,t})\bigr]$. To evaluate how good this loss is, we compare it to a baseline. To this end, we choose a *competitor class* $\mathcal{C}_T$, which is simply a set of deterministic action sequences of length $T$. Intuitively, we would like to compare the player's loss with the cumulative loss of the best action sequence in $\mathcal{C}_T$. In practice, the most common way to evaluate the player's performance is to measure his *external pseudo-regret compared to $\mathcal{C}_T$* (Auer et al., 2002) (which we abbreviate as *regret*), defined as

$$\max_{(y_1,\ldots,y_T)\in \mathcal{C}_T} \mathbb{E}\left[\sum_{t=1}^T f_t(\mathbf{X}_{1,\ldots,t}) - f_t(\mathbf{X}_{1,\ldots,t-1}, y_t)\right] . \quad (1)$$

Most of the theoretical work on online learning uses this definition, both in the bandit setting (e.g., (Auer et al., 2002; Awerbuch & Kleinberg, 2004; Kleinberg, 2004; Flaxman et al., 2005; Bartlett et al., 2008; Abernethy et al., 2008; Hazan & Kale, 2009)) and in the full information setting (e.g., (Zinkevich, 2003; Cesa-Bianchi & Lugosi, 2006; Hazan et al., 2006; Blum & Mansour, 2007; Hazan & Kale, 2009)).

If the adversary is oblivious, regret has a simple and intuitive meaning. In this special case, we can slightly overload our notation and rewrite $f_t(x_1, \ldots, x_t)$ as $f_t(x_t)$. With this simplified notation, the regret defined in Eq. (1) becomes

$$\mathbb{E}\left[\sum_{t=1}^T f_t(X_t)\right] - \min_{(y_1,\ldots,y_T)\in \mathcal{C}_T} \sum_{t=1}^T f_t(y_t) .$$

The above is the difference between the player's loss and the loss of the best sequence in the competitor class $\mathcal{C}_T$. Intuitively, this difference measures how much the player regrets choosing his action-sequence over the best sequence in $\mathcal{C}_T$.

However, if the adversary is adaptive, this simple intuition no longer applies, and the standard notion of regret losses much of its meaning. To observe the problem, note that if the player would have chosen a sequence from the competitor class, say $(y_1, \ldots, y_T)$, then his loss would have been $\sum_{t=1}^T f_t(y_1, \ldots, y_t)$. However, the definition of regret in Eq. (1) instead compares the player's loss to the term $\mathbb{E}\bigl[\sum_{t=1}^T f_t(\mathbf{X}_{1,\ldots,t-1}, y_t)\bigr]$. We can attempt to articulate the meaning of this term: it is the loss in the peculiar situation where the adversary reacts to the player's original sequence $(X_1, \ldots, X_T)$, but the player somehow manages to secretly play the sequence $(y_i, \ldots, y_T)$. This is not a feasible situation and it is unclear why this quantity is an interesting baseline for comparison.

As designers of online learning algorithms, we actually have two different ways to obtain a small regret: we can either design an algorithm that attempts to minimize its loss $\sum_{t=1}^T f_t(\mathbf{X}_{1,\ldots,t})$, or we can cheat by designing an algorithm that attempts to maximize $\sum_{t=1}^T f_t(\mathbf{X}_{1,\ldots,t-1}, y_t)$. For example, consider an algorithm that identifies an action to which the adversary always responds (on the next round) with a loss function that constantly equals 1 (here we use our assumption that the adversary may play any strategy, not necessarily the most malicious one). Repeatedly playing that action would cause regret to asymptote to a constant (the best possible outcome), since the player's loss would grow at an identical rate to the loss of all of its competitors. While this algorithm minimizes regret, it certainly isn't learning how to choose good actions. It is merely learning how to make its competitors look bad.

The problem described above seems to be largely overlooked in the online learning literature, with the exception of two important yet isolated papers (Merhav et al., 2002; de Farias & Megiddo, 2006). To overcome this problem, we define the *policy regret* of the player as the difference between his loss after $T$ rounds and the loss that he would have suffered had he played the best sequence from a competitor class. Policy regret captures the idea that the adversary may react differently to different action sequences. We focus on the bandit setting and start with the class of constant-action competitors. We first prove a negative result: no online bandit algorithm can guarantee a sublinear policy regret. However, if the adversary has a bounded memory, we show how a simple mini-batching technique converts an online bandit algorithm with a regret bound of $O(T^q)$ into an algorithm with a policy regret bound of $O(T^{1/(2-q)})$. We use this technique to derive a policy-regret bound of $O(T^{2/3})$ for the $k$-armed bandit problem, $O(T^{4/5})$ for bandit convex optimization, $O(T^{3/4})$ for bandit linear optimization (or $O(T^{2/3})$ if the player knows the adversary's memory size), and $O(T^{3/4})$ for bandit submodular optimization. We then extend our technique to other notions of regret, namely, switching regret, internal regret, and swap regret.

### 1.1. Related Work

The pioneering work of Merhav et al. (2002) addresses the problem discussed above in the *experts* setting (the full-information version of the $k$-armed bandit problem) and presents a concrete full-information algorithm with a policy regret of $O(T^{2/3})$ against memory-bounded adaptive adversaries. Our work extends and improves on Merhav et al. (2002) in various ways. First, Merhav et al. (2002) are not explicit about the shortcomings of the standard regret definition. Second, note that a bandit algorithm can always be run in the full-information setting (by ignoring the extra feedback) so all of our results also apply to the full-information setting and can be compared to those of Merhav et al. (2002). While Merhav et al. (2002) present one concrete algorithm with a policy regret bound, we show a general technique that endows any existing bandit algorithm with a policy regret bound. Despite the wider scope of our result, our proofs are simpler and shorter than those in Merhav et al. (2002) and our bound is just as good. Our extensions to switching regret, internal regret, and swap regret are also entirely novel.

The work of de Farias & Megiddo (2006) is even more closely related to ours as it presents a family of algorithms that deal with adaptive adversaries in the bandit setting. However, it is difficult to compare the results in de Farias & Megiddo (2006) with our results. While we stick with the widely accepted notion of online regret, de Farias & Megiddo (2006) forsake the notion of regret and instead analyze their algorithms using a non-standard formalization. Moreover, their analysis makes the assumption that the true value of any constant action can be estimated by repeating that action for a sufficiently large number of rounds, at any point in the game.

The reinforcement learning (RL) literature is also related to our work, at least in spirit. Specifically, the PAC-MDP framework (Szepesvári, 2010, section 2.4.2) models the player's *state* on each round; typically, there is a finite number $S$ of states and the player's actions both incur a loss and cause him to transition from one state to another. The PAC-MDP bounds typically hold when the comparison is with all $k^S$ policies (mappings from states to actions), not just the $k$ constant-action policies. Our work is still substantially different from RL. The state transitions in RL are often assumed to be stochastic, whereas our setting is adversarial. An adversarial variant of the MDP setting was studied in Even-Dar et al. (2009), however, it assumes that *all* loss functions across all states are observed by the player. There are recent extensions (Yu et al., 2009; Neu et al., 2010) to the partial feedback or bandit setting but they either give asymptotic rates or make even more stringent assumptions on the underlying state transition dynamics. Also, the dependence on the number of states, $S$, tends to be of the form $\Theta(S^\alpha)$ for some $\alpha > 0$. In the case of an $m$-memory bounded adaptive adversary, associating a state with each possible $m$-length history results in an exponential number of states.

In other related work, Ryabko & Hutter (2008) address the question of learnability in a general adaptive stochastic environment. They prove that environments that allow a rapid recovery from mistakes are asymptotically learnable. At a high level, the assumption that the adaptive adversary is memory bounded serves the same purpose. Our results differ from theirs in various ways: we consider adversarial environments rather than stochastic ones, we present a concrete tractable algorithm whereas their algorithm is intractable, and we prove finite-horizon convergence rates while their analysis is asymptotic.

More recently, Maillard & Munos (2010) considered adaptive adversaries in the $k$-armed bandit setting. They define a framework where the set of all action-histories is partitioned into equivalence classes. For example, assuming that the adversary is $m$-memory-bounded is the same as assuming that two action-histories with a common suffix of length $m$ are equivalent. Within this framework, they study adversaries whose losses are functions of the equivalence classes and competitors whose actions are functions of the equivalence classes. However, they still use the standard notion of regret and do not address its intuitive problems. As mentioned above, this makes their regret bounds difficult to interpret. Moreover, when faced with an $m$-memory-bounded adversary, their bounds and running time both grow exponentially with $m$.

## 2. Policy regret

Define the player's *policy regret* compared to a competitor class $\mathcal{C}_T$ as

$$\mathbb{E}\left[\sum_{t=1}^T f_t(\mathbf{X}_{1,\ldots,t})\right] - \min_{(y_1,\ldots,y_T)\in\mathcal{C}_T} \sum_{t=1}^T f_t(y_1,\ldots,y_t) \ .$$

This coincides with Eq. (1) for oblivious adversaries.

First, we show a negative result. Let $\mathcal{C}_T$ be the set of constant action sequences, namely, sequences of the form $(y,\ldots,y)$ for $y \in \mathcal{X}$. We prove that it is impossible to obtain a non-trivial (sublinear) upper-bound on policy regret that holds for all adaptive adversaries.

**Theorem 1.** *For any player there exists an adaptive adversary such that the player's policy regret compared to the best constant action sequence is $\Omega(T)$.*

*Proof.* Let $y \in \mathcal{X}$ and $p \in (0,1]$ be such that $\Pr(X_1 = y) = p$. Define an adaptive adversary that chooses the loss functions $f_1(x_1) = 0$ and

$$\forall\, t \geq 2 \quad f_t(x_1,\ldots,x_t) = \begin{cases} 1 & \text{if } x_1 = y \\ 0 & \text{if } x_1 \neq y \end{cases} \ .$$

All of the loss functions defined above are constant functions of the current action. From round two and on, the value of the loss function depends entirely on whether the player's first action was $y$ or not. The player's expected cumulative loss against this adversary equals $pT$, since the probability that $X_1 = y$ equals $p$. On the other hand, if the player were to play any constant sequence other than $(y,\ldots,y)$, it would accumulate a loss of zero. Therefore, the player's policy regret is at least $pT$. For comparison, note that the player's (standard) regret is zero. $\square$

Other adversarial strategies can cause specific algorithms to suffer a linear regret. For example, the popular EXP3 algorithm (Auer et al., 2002) for the $k$-armed bandit problem maintains a distribution $(p_{1,t},\ldots,p_{k,t})$ over the $k$ arms on each round. This distribution is a deterministic function of the algorithm's past observations. If the adversary mimics EXP3's computations and sets the loss to be $f_t(j) = p_{j,t}$ we can prove that this distribution converges to the uniform distribution and EXP3 suffers a linear loss. In contrast, playing any constant arm against this adversary results in a sublinear loss, which implies a linear policy regret.

Given that no algorithm can guarantee a small policy regret against all adaptive adversaries, we must restrict the set of possible adversaries. We consider an adversary that lies between oblivious and adaptive: An *$m$-memory-bounded adaptive adversary* is an adversary that is constrained to choose loss functions that depend only on the $m+1$ most recent actions. Formally,

$$f_t(x_1,\ldots,x_t) \;=\; f_t(x'_1,\ldots,x'_{t-m-1},x_{t-m},\ldots,x_t) \ ,$$

for all $x_1,\ldots,x_t$ and $x'_1,\ldots,x'_{t-m-1}$ in $\mathcal{X}$. An oblivious adversary is 0-memory-bounded, while a general adaptive adversary is $\infty$-memory-bounded. We note that $m$-memory-bounded adversaries arrise in many natural scenarios. For example, the friction cost associated with switching from one action to another can be modeled using a 1-memory-bounded adversary.

For $m$-memory-bounded adaptive adversaries we prove a positive result, in the form of a reduction. Again, let the competitor class $\mathcal{C}_T$ be the set of all constant action sequences of length $T$. We show how an algorithm $\mathcal{A}$ with a sublinear (standard) regret bound against an adaptive adversary can be transformed into another algorithm with a (slightly-inferior) policy regret bound against an $m$-memory-bounded adaptive adversary. We note that this new algorithm does not need to know $m$, but $m$ does appear as a constant in our analysis.

We define a new algorithm by wrapping $\mathcal{A}$ with a mini-batching loop (e.g., Dekel et al. (2011)). We specify a batch size $\tau$ and name the new algorithm $\mathcal{A}_\tau$. The algorithm $\mathcal{A}_\tau$ groups the online rounds $1,\ldots,T$ into consecutive and disjoint mini-batches of size $\tau$: The $j$'th mini-batch begins on round $(j-1)\tau+1$ and ends on round $j\tau$. At the beginning of mini-batch $j$, $\mathcal{A}_\tau$ invokes $\mathcal{A}$ and receives an action $Z_j$ drawn from $\mathcal{A}$'s internal distribution over the action space. Then, $\mathcal{A}_\tau$ plays this action for $\tau$ rounds, namely, $X_{(j-1)\tau+1} = \ldots = X_{j\tau} = Z_j$. During the mini-batch, $\mathcal{A}$ does not observe any feedback, does not update its internal state, and is generally unaware that $\tau$ rounds are going by. At the end of the mini-batch, $\mathcal{A}_\tau$ feeds $\mathcal{A}$ with a single loss value, the average loss suffered during the mini-batch, $\frac{1}{\tau}\sum_{t=(j-1)\tau+1}^{j\tau} f_t(\mathbf{X}_{1,\ldots,t})$.

From $\mathcal{A}$'s point of view, every mini-batch feels like a single round: it chooses a single action $Z_j$, receives a single loss value as feedback, and updates its internal state once. Put more formally, $\mathcal{A}$ is performing standard online learning with bandit feedback against the loss sequence $\hat{f}_1, \ldots, \hat{f}_J$, where $J = \lfloor T/\tau \rfloor$,

$$\hat{f}_j(z_1, \ldots, z_j) = \frac{1}{\tau} \sum_{k=1}^{\tau} f_{(j-1)\tau+k}(\mathbf{z}_1^\tau, \ldots, \mathbf{z}_{j-1}^\tau, \mathbf{z}_j^k), \quad (2)$$

and $\mathbf{z}^i$ denotes $i$ repetitions of the action $z$. By assumption, $\mathcal{A}$'s regret against $\hat{f}_1, \ldots, \hat{f}_J$ is upper bounded by a sublinear function of $J$. The following theorem transforms this bound into a bound on the policy regret of $\mathcal{A}_\tau$.

**Theorem 2.** *Let $\mathcal{A}$ be an algorithm whose (standard) regret, compared to constant actions, against any sequence of $J$ loss functions generated by an adaptive adversary, is upper bounded by a monotonic function $R(J)$. Let $\tau > 0$ be a mini-batch size and let $\mathcal{A}_\tau$ be the mini-batched version of $\mathcal{A}$. Let $(f_t)_{t=1}^T$ be a sequence of loss functions generated by an $m$-memory-bounded adaptive adversary, let $X_1, \ldots, X_T$ be the sequence of actions played by $\mathcal{A}_\tau$ against this sequence, and let $y$ be any action in $\mathcal{X}$. If $\tau > m$, the policy regret of $\mathcal{A}_\tau$, compared to the constant action $y$, is bounded by*

$$\mathbb{E}\left[\sum_{t=1}^T f_t(X_{1,\ldots,t}) - f_t(\mathbf{y}^t)\right] \le \tau R\left(\frac{T}{\tau}\right) + \frac{Tm}{\tau} + \tau.$$

*Specifically, if $R(J) = CJ^q + o(J^q)$ for some $C > 0$ and $q \in (0,1)$, and $\tau$ is set to $C^{\frac{-1}{2-q}} T^{\frac{1-q}{2-q}}$, then*

$$\mathbb{E}\left[\sum_{t=1}^T f_t(X_{1,\ldots,t}) - f_t(\mathbf{y}^t)\right] \le C' T^{\frac{1}{2-q}} + o(T^{\frac{1}{2-q}}),$$

*where $C' = (m+1)C^{\frac{1}{2-q}}$.*

*Proof.* Assume that $\tau > m$, otherwise the theorem makes no claim. Let $J = \lfloor T/\tau \rfloor$ and let $(\hat{f}_j)_{j=1}^J$ be the sequence of loss functions defined in Eq. (2). Let $Z_1, \ldots, Z_{J+1}$ be the sequence of actions played by $\mathcal{A}$ against the loss sequence $(\hat{f}_j)_{j=1}^J$. Our assumption on $\mathcal{A}$ implies that

$$\mathbb{E}\left[\sum_{j=1}^J \hat{f}_j(\mathbf{Z}_{1,\ldots,j}) - \hat{f}_j(\mathbf{Z}_{1,\ldots,j-1}, y)\right] \le R(J). \quad (3)$$

From the definitions of $\mathcal{A}_\tau$ and $\hat{f}_j$,

$$\sum_{j=1}^J \hat{f}_j(\mathbf{Z}_{1,\ldots,j}) = \frac{1}{\tau} \sum_{t=1}^{J\tau} f_t(\mathbf{X}_{1,\ldots,t}). \quad (4)$$

Introducing the notation $t_j = (j-1)\tau$, we rewrite

$$\sum_{j=1}^J \hat{f}_j(\mathbf{Z}_{1,\ldots,j-1}, y) = \frac{1}{\tau} \sum_{j=1}^J \sum_{k=1}^\tau f_{t_j+k}(\mathbf{X}_{1,\ldots,t_j}, \mathbf{y}^k). \quad (5)$$

For any $j \le J$, the bound on the loss implies

$$\sum_{k=1}^m \left(f_{t_j+k}(\mathbf{X}_{1,\ldots,t_j}, \mathbf{y}^k) - f_{t_j+k}(\mathbf{y}^{t_j+k})\right) \le m, \quad (6)$$

and our assumption that the adversary is $m$-memory-bounded implies

$$\sum_{k=m+1}^\tau f_{t_j+k}(\mathbf{X}_{1,\ldots,t_j}, \mathbf{y}^k) = \sum_{k=m+1}^\tau f_{t_j+k}(\mathbf{y}^{t_j+k}). \quad (7)$$

Combining Eqs. (5-7) gives the bound

$$\sum_{j=1}^J \hat{f}_j(\mathbf{Z}_{1,\ldots,j-1}, y) \le \frac{1}{\tau}\left(\sum_{t=1}^{J\tau} f_t(\mathbf{y}^t) + Jm\right).$$

Together with Eq. (3) and Eq. (4), we have

$$\mathbb{E}\left[\sum_{t=1}^{J\tau} f_t(\mathbf{X}_{1,\ldots,t}) - f_t(\mathbf{y}^t)\right] \le \tau R(J) + Jm.$$

We can bound the regret on rounds $J\tau+1, \ldots, T$ by $\tau$. Plugging in $J \le T/\tau$ gives an overall policy regret bound of $\tau R(T/\tau) + Tm/\tau + \tau$. Focusing on the special case where $R(J) = CJ^q + o(J^q)$, the bound becomes

$$CT^q \tau^{1-q} + Tm\tau^{-1} + \tau + o\left(T^q \tau^{1-q}\right).$$

Plugging in $\tau = C^{\frac{-1}{2-q}} T^{\frac{1-q}{2-q}}$ concludes the proof. □

## 3. Applying the Result

With Thm. 2 in hand, we prove that the policy regret of existing online bandit algorithms grows sublinearly with $T$. We begin with the EXP3 algorithm (Auer et al., 2002) in the classic $k$-armed bandit setting, with its regret bound of $\sqrt{7Jk \log k}$ against any sequence of $J$ loss functions generated by an adaptive adversary. Applying Thm. 2 with $C = \sqrt{7k \log k}$ and $q = 1/2$ proves the following result.

**Corollary 1** ($k$-armed bandit). *Let $\mathcal{X} = \{1, \ldots, k\}$ and let $\mathcal{F}_t$ consist of all functions from $\mathcal{X}^t$ to $[0,1]$. The policy regret of the mini-batched version of the EXP3 algorithm (Auer et al., 2002), with batch size $\tau = (7k \log k)^{-1/3} T^{1/3}$, against an $m$-memory bounded adaptive adversary, is upper bounded by*

$$(m+1)(7k \log k)^{1/3} T^{2/3} + o(T^{2/3}).$$

We move on to the bandit convex optimization problem. The algorithm and analysis in Flaxman et al. (2005) guarantees a regret bound of $18d(\sqrt{LD}+1)J^{3/4}$ against any sequence of $J$ loss functions generated by an adaptive adversary, where $d$ is the dimension, $D$ is the diameter of $\mathcal{X}$, and $L$ is the Lipschitz coefficient of the loss functions. Applying Thm. 2 with $C = 18d(\sqrt{LD}+1)$ and $q = 3/4$ proves the following result.

**Corollary 2** (Bandit convex optimization). *Let $\mathcal{X} \subset \mathbb{R}^d$ be a closed bounded convex set with diameter $D$ and let $\mathcal{F}_t$ be the class of functions from $\mathcal{X}^t$ to $[0,1]$ that are convex and $L$-Lipschitz in their last argument. The policy regret of the mini-batched version of Flaxman, Kalai, and McMahan's algorithm (Flaxman et al., 2005), with batch size $\tau = (18d(\sqrt{LD}+1))^{-4/5}T^{1/5}$, against an $m$-memory bounded adaptive adversary is upper bounded by*

$$(m+1)\bigl(18d(\sqrt{LD}+1)T\bigr)^{4/5} + o(T^{4/5}).$$

An important special case of bandit convex optimization is bandit linear optimization. The analysis in Dani & Hayes (2006) proves a regret bound of $15dJ^{2/3}$ for the algorithm of McMahan & Blum (2004) against any sequence of $J$ loss functions generated by an adaptive adversary. Applying Thm. 2 with $C = 15d$ and $q = 2/3$ proves the following result.

**Corollary 3** (Bandit linear optimization). *Let $\mathcal{X} \subset [-2,2]^d$ be a polytope (or, more generally, let $\mathcal{X} \subseteq \mathbb{R}^d$ be a convex set over which linear optimization can be done efficiently). Let $\mathcal{F}_t$ be the class of functions from $\mathcal{X}^t$ to $[0,1]$ that are linear in their last argument. The policy regret of the mini-batched version of McMahan and Blum's algorithm (McMahan & Blum, 2004), with batch size $\tau = (15d)^{-3/4}T^{1/4}$, against an $m$-memory bounded adaptive adversary is upper bounded by*

$$(m+1)(15dT)^{3/4} + o(T^{3/4}).$$

Finally, we apply our result to bandit submodular minimization over a ground set $\{1, \ldots, k\}$. Recall that a set function $f$ is submodular if for any two subsets of the ground set $A, B$ it holds that $f(A \cup B) + f(A \cap B) \leq f(A) + f(B)$. The algorithm in Hazan & Kale (2009) has a regret bound of $12kJ^{2/3}$ against any sequence of $J$ loss functions generated by an adaptive adversary. Applying Thm. 2 with $C = 12k$ and $q = 2/3$ proves the following result.

**Corollary 4** (Bandit submodular minimization). *Let $\mathcal{X}$ be the power set of $\{1, \ldots, k\}$ and let $\mathcal{F}_t$ be the class of functions from $\mathcal{X}^t$ to $[0,1]$ that are submodular in their last argument. The policy regret of the mini-batched version of Hazan and Kale's algorithm (Hazan*

*& Kale, 2009), with batch size $\tau = (12k)^{-3/4}T^{1/4}$, against an $m$-memory bounded adaptive adversary is*

$$(m+1)(12kT)^{3/4} + o(T^{3/4}).$$

## 4. Extensions

Theorem 2 is presented in its simplest form, and we can extend it in various interesting ways.

### 4.1. Relaxing the Adaptive Assumption

Recall that we assumed that $\mathcal{A}$ has a (standard) regret bound that holds for any loss sequence generated by an adaptive adversary. A closer look at the proof of Thm. 2 reveals that it suffices to assume that $\mathcal{A}$'s regret bound holds against any loss sequence generated by a 1-memory-bounded adaptive adversary. To see why, note that the assumption that each $f_t$ is $m$-memory-bounded, the assumption that $\tau > m$, and the definition of $\hat{f}_j$ in Eq. (2) together imply that each $\hat{f}_j$ is 1-memory-bounded.

### 4.2. When $m$ is Known

We can strengthen Thm. 2 in two ways if the memory bound $m$ is given to $\mathcal{A}_\tau$. First, we redefine $\hat{f}_j(z_1, \ldots, z_j)$ as

$$\frac{1}{\tau - m} \sum_{k=m+1}^{\tau} f_{(j-1)\tau + k}(\mathbf{z}_1^\tau, \ldots, \mathbf{z}_{j-1}^\tau, \mathbf{z}_j^k). \quad (8)$$

Note that the first $m$ rounds in the mini-batch are omitted. This makes the sequence $(\hat{f}_j)_{j=1}^J$ a 0-memory-bounded sequence. In other words, we only need $\mathcal{A}$'s regret bound to hold for *oblivious* adversaries. In addition to relaxing the assumption on the regret of the original algorithm $\mathcal{A}$, we use $m$ to further optimize the value of $\tau$. This reduces the linear dependence on $m$ in our policy regret bounds, as seen in the following example. We focus, once again, on the bandit linear optimization setting. We use the algorithm of Abernethy, Hazan, and Rakhlin (Abernethy et al., 2008), whose regret bound is $16n\sqrt{\vartheta J \log J}$ when $J > 8\vartheta \log J$, for any sequence of $J$ loss functions generated by an *oblivious* adversary. The constant $\vartheta$ is associated with a self-concordant barrier on $\mathcal{X}$. In the current context, understanding the nature of this constant is unimportant, and it suffices to know that $\vartheta = O(n)$ when $\mathcal{X}$ is a closed convex set (Nesterov & Nemirovsky, 1994).

**Theorem 3** (Bandit linear optimization, known $m$). *Let $\mathcal{X}$ be a convex set and let $\mathcal{F}_t$ be the class of functions from $\mathcal{X}^t$ to $[0,1]$ that are linear in their last argument. In this setting, The policy regret of the mini-batched version of Abernethy, Hazan, and Rakhlin's* algorithm (Abernethy et al., 2008) where the first $m$ loss values in each mini-batch are ignored, with batch size $\tau = m^{2/3}(16n\sqrt{\vartheta}\log T)^{-2/3}T^{1/3}$, against an $m$-memory-bounded adaptive adversary, is upper bounded for all $T > 8\vartheta \log T$ by

$$2m^{1/3}(16n\sqrt{\vartheta}T\log T)^{2/3} + O(T^{1/3}).$$

### 4.3. Switching Competitors

So far, we defined $\mathcal{C}_T$ to be the simplest competitor class possible, the class of constant action sequences. We now redefine $\mathcal{C}_T$ to include all piece-wise constant sequences with at most $s$ switches (Auer et al., 2002). Namely, a sequence in $\mathcal{C}_T$ is a concatenation of at most $s$ shorter constant sequences, whose total length is $T$. In this case, we assume that $\mathcal{A}$'s regret bound holds compared to sequences with $s$ switches and we obtain a policy regret bound that holds compared to sequences with $s$ switches.

**Theorem 4.** *Repeat the assumptions of Thm. 2, except that $\mathcal{C}_T$ is the set of action sequences with at most $s$ switches (where $s$ is fixed and independent of $T$) and $\mathcal{A}$'s regret bound of $R(J)$ holds compared to action-sequences in $\mathcal{C}_J$. Then, the policy regret of $\mathcal{A}_\tau$, compared to action-sequences in $\mathcal{C}_T$, against the loss sequence $(f_t)_{t=1}^T$, is upper bounded by*

$$\tau R(T/\tau) + Tm/\tau + (s+1)\tau.$$

The main observation required to prove this lemma is that our proof of Thm. 2 bounds the regret batch-by-batch. The $s$ switches of the competitor's sequence may affect at most $s$ batches. We can trivially upper bound the regret on these batches using the fact that the loss is bounded, adding $s\tau$ to the overall bound.

In the $k$-armed bandit setting, (Auer et al., 2002) defines an algorithm named EXP3.S and proves a regret bound compared to sequences with $s$ switches. Combining the guarantee of EXP3.S with the lemma above gives the following result.

**Theorem 5** ($k$-armed bandit with switches). *Let $\mathcal{X} = \{1,\ldots,k\}$ and let $\mathcal{F}_t$ consist of all functions from $\mathcal{X}^t$ to $[0,1]$. The policy regret of the mini-batched version of the EXP3.S algorithm, with batch size $\tau = (7ks\log(kT))^{-1/3}T^{1/3}$, compared to action sequences with at most $s$ switches, against an $m$-memory bounded adaptive adversary, is upper bounded by*

$$(m+1)\big(7ks\log(kT)\big)^{1/3}T^{2/3} + O(T^{1/3}).$$

It is possible to give similar guarantees for settings such as bandit convex optimization, provided that regret guarantees under action switches are available.

For instance, Flaxman et al. (Flaxman et al., 2005, Section 4) talk about (but do not explicitly derive) extensions of their bandit convex optimization regret guarantees that incorporate switches.

### 4.4. Internal Regret, Swap Regret, $\Phi$-Regret

We have so far considered the standard notion of external pseudo-regret, where the player's action-sequence is compared to action sequences in a class $\mathcal{C}_T$, where $\mathcal{C}_T$ is commonly chosen to be the set of constant sequences. Other standard (yet less common) ways to analyze the performance of the player use the notions of *internal regret* (Blum & Mansour, 2007) and *swap regret* (Blum & Mansour, 2007). To define these notions, let $\Phi$ be a set of action transformations, namely each $\phi \in \Phi$ is a function of the form $\phi : \mathcal{X} \to \mathcal{X}$. The player's $\Phi$-regret is then defined as:

$$\max_{\phi \in \Phi} \mathbb{E}\left[\sum_{t=1}^T f_t(X_{1,\ldots,t}) - f_t(X_{1,\ldots,t-1}, \phi(X_t))\right].$$

In words, we compare the player's loss to the loss that would have been attained if the player had replaced his current action according to one of the transformations in $\Phi$. We recover external regret compared to constant action sequences by letting $\Phi$ be the set of constant functions, that map all actions to a constant action $y$. Internal regret is defined by setting $\Phi = \{\phi_{y \to y'} : y, y' \in \mathcal{X}\}$, where

$$\phi_{y \to y'}(x) = \begin{cases} y' & \text{if } x = y \\ x & \text{otherwise} \end{cases}.$$

In other words, $\phi_{y \to y'}$ replaces all occurrences of action $y$ with action $y'$, but leaves all other actions unmodified. To define swap regret, we specialize to the $k$-armed bandit case, where $\mathcal{X} = \{1,\ldots,k\}$. Swap regret is defined by setting $\Phi = \{\phi_{y_1,\ldots,y_k} : \forall j\ y_j \in \mathcal{X}\}$, where $\phi_{y_1,\ldots,y_k}(x) = y_x$. In other words, this function replaces every action with a different action.

$\Phi$-regret suffers from the same intuitive difficulty as external regret, when the adversary is adaptive. Define the *policy $\Phi$-regret* as

$$\max_{\phi \in \Phi} \mathbb{E}\left[\sum_{t=1}^T f_t(X_{1,\ldots,t}) - f_t\big(\phi(X_1),\ldots,\phi(X_t)\big)\right].$$

We repeat our technique to prove the following.

**Theorem 6.** *Repeat the assumptions of Thm. 2, except that now let $\Phi$ be any set of action transformations and assume that $\mathcal{A}$'s $\Phi$-regret against any sequence of $J$ loss functions generated by an adaptive* adversary is upper bounded by $R(J)$. Then, the policy $\Phi$-regret of $\mathcal{A}_\tau$ against $(f_t)_{t=1}^T$ generated by an $m$-memory-bounded adaptive adversary is bounded by $\tau R(T/\tau) + Tm/\tau + \tau$.

The proof is omitted due to space constraints.

Blum & Mansour (2007) presents a technique of converting any online learning algorithm with an external regret bound into an algorithm with an internal regret bound. Combining that technique with ours endows any of the online learning algorithms mentioned in this paper with a bound on internal policy regret.

## 5. Discussion

We highlighted a problem with the standard definition of regret when facing an adaptive adversary. We defined the notion of policy regret and argued that it captures the intuitive semantics of the word "regret" better than the standard definition. We then went ahead to prove non-trivial upper bounds on the policy regret of various bandit algorithms.

The main gap in our current understanding of policy regret is the absence of lower bounds (in both the bandit and the full-information settings). In other words, we do not know how tight our upper bounds are. It is conceivable that bandit algorithms that are specifically designed to minimize policy regret will have superior bounds, but we are yet unable to show this. On a related issue, we do not know if our mini-batching technique is really necessary: perhaps one could prove a non-trivial policy regret bound for the original (unmodified) EXP3 algorithm. We leave these questions as open problems for future research.